\pdfoutput=1

\documentclass[11pt]{article}

\usepackage[preprint]{acl}
\usepackage{times}
\usepackage{latexsym}
\usepackage[T1]{fontenc}

\usepackage[utf8]{inputenc}

\usepackage{microtype}

\usepackage{inconsolata}

\usepackage{graphicx}

\title{NormTab: Improving Symbolic Reasoning in LLMs Through Tabular Data Normalization}

\author{Md Mahadi Hasan Nahid \\ University of Alberta \\mnahid@ualberta.ca
        \And  
        Davood Rafiei \\ University of Alberta \\ drafiei@ualberta.ca}

\begin{document}
\maketitle
\begin{abstract}

In recent years, Large Language Models (LLMs) have demonstrated remarkable capabilities in parsing textual data and generating code. However, their performance in tasks involving tabular data, especially those requiring symbolic reasoning, faces challenges due to the structural variance and inconsistency in table cell values often found in web tables. In this paper, we introduce NormTab, a novel framework aimed at enhancing the symbolic reasoning performance of LLMs by normalizing web tables. We study table normalization as a stand-alone, one-time preprocessing step using LLMs to support symbolic reasoning on tabular data. Our experimental evaluation, conducted on challenging web table datasets such as WikiTableQuestion and TabFact, demonstrates that leveraging NormTab significantly improves symbolic reasoning performance, showcasing the importance and effectiveness of web table normalization for enhancing LLM-based symbolic reasoning tasks.


\end{abstract}

\section{Introduction}

Tables are a fundamental format for structured data representation and are widely used across various sources, including relational databases, web pages, and financial documents. However, many tables within documents and web pages are designed for direct human consumption and often lack the strict formatting that is expected in relational tables. This discrepancy poses significant challenges when querying them using languages such as SQL, integrating them with relational databases, and processing them within applications.

Large Language Models (LLMs) \citep{brown2020language} have emerged as powerful tools for semantic parsing both textual and tabular data and performing complex tasks such as code generation. Trained on vast amount of Internet data,  including both text and tables, and employing techniques such as Chain of Thought (CoT) prompting \citep{wei2022chain} and self-consistency \citep{wang2023selfconsistency}, these models outperform many traditional models on various table reasoning tasks~\citep{gu-etal-2022-pasta, Chen2020TabFact:, herzig-etal-2020-tapas, wang-etal-2019-learning}.
However, their performance in tasks involving tabular data, particularly those requiring symbolic reasoning, is often hindered by the structural variability and inconsistencies commonly found in web tables. Symbolic reasoning over tables necessitates a clear understanding of the table structure and values, and may involve constraining rows and columns, which can be challenging when dealing with unstructured or noisy web tables \citep{pourreza2023dinsql, ni2023lever, cheng2022binding, zhang2023reactable}.
Our hypothesis is that normalizing ill-formatted tables can address this challenge, enabling the execution of symbolic programs (such as SQL or Python) on the tables and making reasoning tasks involving comparison, aggregation, and mathematical calculations more manageable. Moreover, normalization may enhance the explainability by allowing the tracking of the intermediate steps in reasoning.



\begin{figure*}[t]
    \centering
    \resizebox{\textwidth}{!}{
    \includegraphics{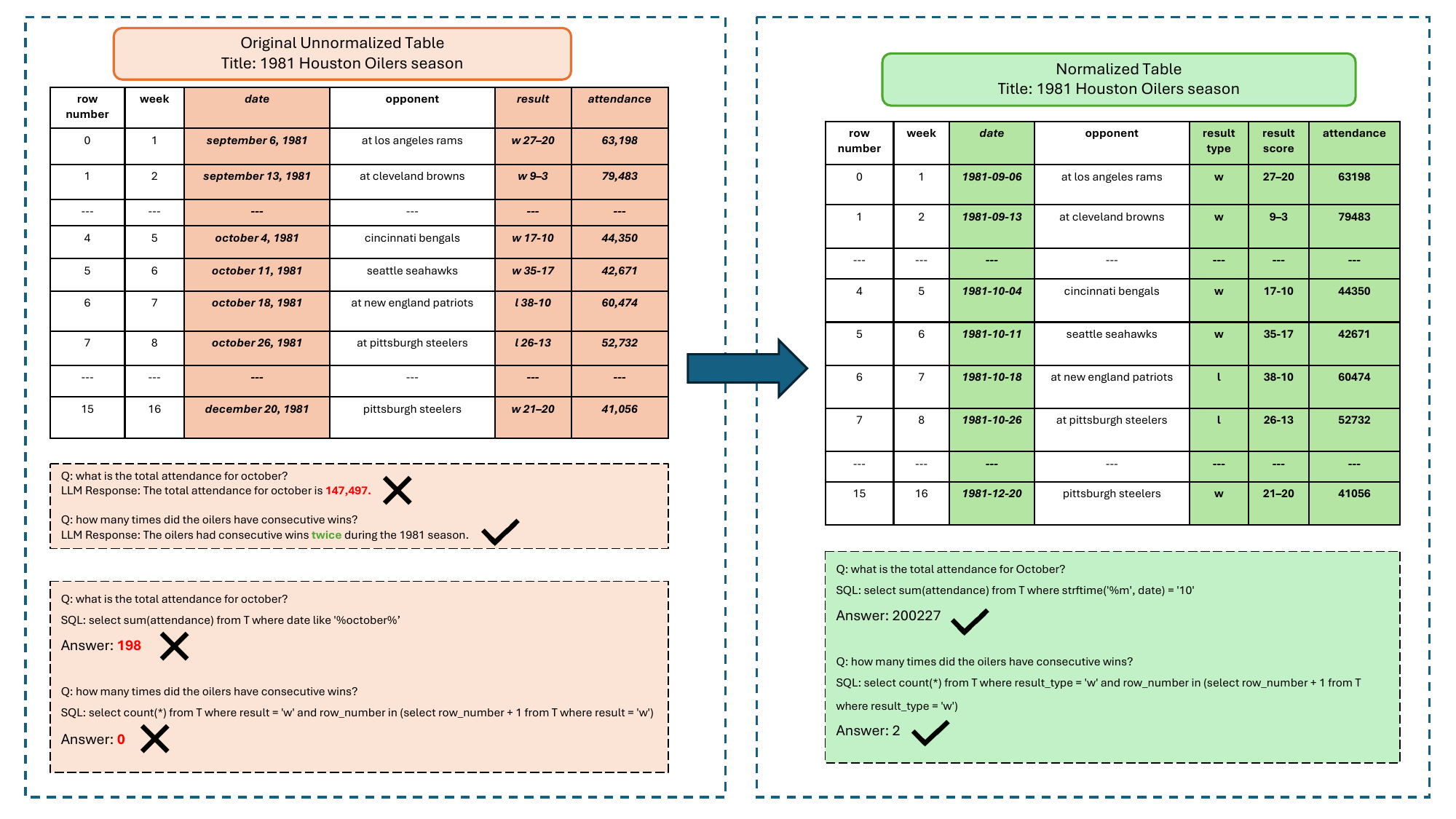}
    }
    \caption{An example of a Table QA task, with the original unnormalized web table shown on the left and its normalized version on the right. Retrieve answers using a symbolic approach from the unnormalized table poses difficulties due to inconsistent formatting of \textit{\textbf{date}}, \textit{\textbf{result}} and \textit{\textbf{attendance}} columns. Also, direct querying with LLMs often fails for questions involving numerical operations. Normalization enables effective text-to-SQL conversion, as shown by the normalized table on the right.}
    \label{fig:NormTab-example}
\end{figure*}

Consider the table QA task shown in Figure~\ref{fig:NormTab-example}. Retrieving answers from the table on the left using a symbolic approach such as SQL is challenging due to the irregular structure of the data and the limitations of SQL. While an LLM may handle simple look-up questions, it struggles with tasks requiring complex aggregation and arithmetic operations. However, the normalized version of the same table, shown on the right, can be easily analyzed, allowing text-to-SQL approaches to effectively obtain the answers to questions.


Existing models for table reasoning typically rely on a multi-step framework, where an LLM performs a sequence of actions such as  adding columns before  additional scripts are invoked to process data, retrieve cell values, or compute answers to questions \citep{liu2023rethinking, wang2024chain, zhang2023reactable}. These models are often dependent on question and table structure and do not address the root cause of table irregularity, making them less scalable.
An alternative is normalizing tables, often part of a larger process known as data wrangling, which involves processing, cleaning and organizing data into a format that is suitable for further analysis. Significant progress has been made on data wrangling~\citep{furche2016data,abedjan2016detecting,rattenbury2017principles}, with recent approaches employing LLMs for tasks such as error detection and data imputation~\citep{narayan2022can}. Selected operations, such as normalizing numbers and dates, may also be introduced into data processing pipelines to facilitate further analysis~\citep{nahid2024tabsqlify}. To the best of our knowledge, our work is the first to study table normalization as an stand-alone one-time preprocessing step using LLMs.


In this paper, we introduce \textbf{NormTab}, a framework designed to normalize web tables to align them with the structured format of relational database tables. NormTab addresses challenges such as structural variance, mixed data formats, and extraneous information, thereby facilitating accurate and efficient symbolic reasoning and query processing using LLMs. Our work explores two key research questions: 

\begin{itemize}
    \item \textbf{RQ1:} How can we leverage LLMs' textual understanding to effectively clean and normalize web tables?
    \item \textbf{RQ2:} How can web table normalization enhance table reasoning tasks, particularly in the context of LLM-based symbolic reasoning? 
\end{itemize}

Our proposed solution leverages the advanced textual understanding capabilities of LLMs to independently process and normalize web tables, without relying on specific questions. By normalizing tables in this manner, we enable a robust foundation for any downstream task involving table reasoning. This approach allows for multiple questions to be asked from a single, normalized table, significantly enhancing reasoning and query capabilities. Moreover, our normalization process only needs to be performed once, unlike other models that require repeated adjustments based on different questions, highlighting a key advantage of our approach.



Through a comprehensive experimental evaluation conducted on challenging web table datasets such as WikiTableQuestions \citep{pasupat-liang-2015-compositional} and TabFact \citep{Chen2020TabFact:}, we assess the effectiveness of NormTab in improving table reasoning performance. These datasets provide diverse examples of table structures and content, allowing us to thoroughly investigate the impact of web table normalization on LLM-based symbolic reasoning tasks. By addressing RQ1 and RQ2, we aim to demonstrate the importance of web table normalization and its potential to enhance the capabilities of LLMs in handling tabular data for complex reasoning tasks.

Key Contributions of our paper are:
\begin{itemize}
    \item We introduce NormTab, a novel framework that enhances LLMs' symbolic reasoning on tabular data by normalizing web tables. NormTab includes structure normalization (e.g., transposing tables, flattening rows and columns) and value normalization (e.g., removing extraneous strings, standardizing the formatting of dates and numbers) to ensure consistency and accuracy in reasoning tasks. 
    \item We demonstrate how LLMs' textual understanding can be effectively utilized for data cleaning and transformation tasks, addressing challenges such as structural variance, mixed values, noise, and substring extraction in web tables 
    \item We conduct extensive experimental evaluations using challenging web table datasets, including WikiTableQuestion and TabFact, to assess the effectiveness of NormTab in improving table reasoning performance, particularly in the context of LLM-based symbolic reasoning tasks.
    
\end{itemize}

\begin{figure*}[t]
    \centering
    \resizebox{\textwidth}{!}{
    \includegraphics{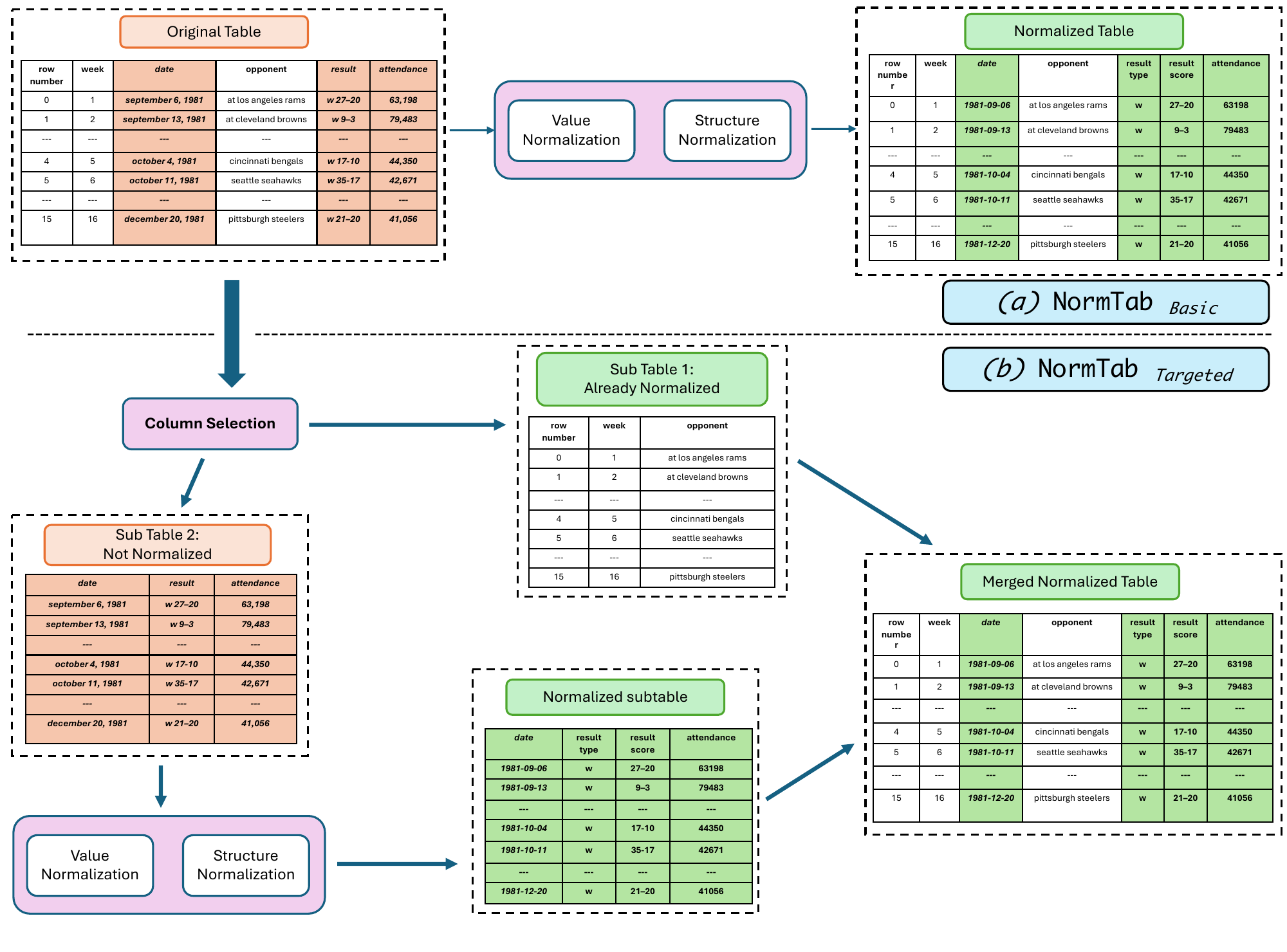}
    }
    \caption{Overview of \textit{\textbf{NormTab}}. The methodology encompasses two distinct strategies: \textit{\textbf{(a) Entire Table Normalization (NormTab\textsubscript{Basic})}:} we provide the LLM with the entire web table along with specific instructions for cleaning and normalizing. The LLM reads the table and the instructions, then returns a cleaned and normalized version of the table. \textit{\textbf{(b) Targeted Normalization (NormTab\textsubscript{Targeted}):}} In this approach the LLM identifies and targets only the portions of the web table requiring normalization based on the table metadata and a few sample rows. The original table is split into two subtables: one for normalization and one already clean. The LLM processes the subtable that requires normalization then returned a cleaned version. Finally, the normalized subtable is merged with the clean portion, resulting in a fully cleaned and normalized table.}
    \label{fig:NormTab}
\end{figure*}

\section{Related Work}
Our work is related to a few areas as discussed next.

\textbf{General LLMs and CoT}
Related to our work is the line of research aimed at improving the performance of LLMs \cite{brown2020language} on various  reasoning tasks, with capabilities spanning mathematics, common sense, and symbolic reasoning~\citep{chen-2023-large, 10.1145/3539618.3591708, cheng2022binding}. 
These approaches often excel using few-shot prompts without requiring fine-tuning. Their reasoning abilities can be further enhanced by breaking complex tasks into steps, employing methods like chain-of-thought (CoT) \cite{wei2022chain} prompting and Zero-CoT. For instance, the Table-CoT \citep{chen-2023-large} model utilizes in-context learning and CoT prompting to generate answers for table-based tasks. 

Several studies have utilized instruction tuning and supervised fine-tuning to enhance the performance of LLMs on table reasoning tasks. Notable examples include TableLLaMA \citep{zhang2023tablellama} and TableGPT \citep{zha2023tablegpt}, which have shown significant improvements in specific applications. In contrast, the BINDER model \citep{cheng2022binding} extends the capabilities of LLMs to programming language generation for solving commonsense problems. Additionally, the DATER approach \citep{10.1145/3539618.3591708} employs LLMs to decompose tables and questions, facilitating table-based QA and fact verification tasks. These diverse approaches underscore the potential of LLMs in handling complex reasoning tasks involving tabular data.\\ 




\textbf{Reasoning over structured data/tables }
Another line of related work is reasoning over tabular data. 
Several studies leverage symbolic reasoning through text-to-SQL or Python code for table-based reasoning tasks. However, for effectively utilizing the symbolic code generation approach with LLMs for table reasoning tasks, it is crucial to ensure that the table is in the proper format~\citep{pourreza2023dinsql, rajkumar2022evaluating, ni2023lever, nahid2024tabsqlify, cheng2022binding}. 

Chain-of-Table \citep{wang2024chain} enhances reasoning on tabular data by iteratively transforming and evolving table structures through a series of reasoning steps, including row/column selection, cell splitting to refine table representations for specific reasoning tasks. Their method employs in-context learning to direct LLMs in iteratively generating operations and updating the table, thus forming a chain of reasoning specific to tabular data. \citet{liu2023rethinking} explore the capabilities of LLMs in interpreting and reasoning over tabular data, emphasizing robustness to structural perturbations, comparing textual and symbolic reasoning, and examining the potential of aggregating multiple reasoning pathways. Their findings indicate that structural variations in tables presenting the same content can significantly degrade performance, particularly in symbolic reasoning tasks. They propose a method for table structure normalization through transposition to mitigate this issue and find that while textual reasoning slightly outperforms symbolic reasoning, each approach has distinct strengths depending on the task.

StructGPT \citep{jiang-etal-2023-structgpt} employs an iterative reading-then-reasoning approach to enhance LLM reasoning for structured data, but its scalability is constrained by token limits when processing large tables. The ReAcTable model \citep{zhang2023reactable} adopts the ReAct paradigm, integrating step-by-step reasoning, external tool-based code execution, intermediate table generation, and majority voting to process tabular data. Similarly, the LEVER model \cite{ni2023lever} improves language-to-code generation by validating generated programs based on their execution results, enhancing the accuracy and reliability of table reasoning tasks.

\textbf{Data wrangling and imputation} Normalizing tables is a crucial aspect of the broader data wrangling process, which involves processing, cleaning, and organizing data into a format suitable for further analysis. Considerable research has focused on data wrangling, addressing challenges such as error detection, data imputation, and standardization of data formats ~\citep{furche2016data,abedjan2016detecting,rattenbury2017principles}. Recent approaches have leveraged the capabilities of LLMs for these tasks. For instance, \citet{narayan2022can} demonstrated the effectiveness of LLMs in identifying errors and imputing missing data, showcasing how these models can enhance the data wrangling process. By integrating LLMs, the efficiency and accuracy of preparing data for analysis can be significantly improved, streamlining and automating many aspects of data wrangling. Operations like normalizing numbers and dates can be incorporated into data processing workflows to aid in subsequent analysis~\citep{nahid2024tabsqlify}. 

All these works highlight the importance of table normalization in improving LLMs' performance on tabular data, paving the way for more effective and accurate table reasoning models. 




\section{Methodology}
Our methodology encompasses several essential parts designed to ready web tables for proficient reasoning by LLMs. 


\subsection{Normalization Operations}

The normalization operations in \textbf{NormTab} can be divided into two groups: \textbf{(1)} value normalization and \textbf{(2)} structural normalization. The former involves splitting cells to add new columns, handling empty cells and value ranges, removing extraneous strings, and normalizing data formats such as dates and numerical values to ensure consistency and accuracy in reasoning tasks. Structural normalization, on the other hand, aims to detect structural variance by analyzing the first row and first column of a web table and determining whether a transposition is needed. If transposition is required, we  address this issue by flipping the rows and columns. \\

\textbf{Value Normalization:} 
Our value normalization is based on the principle that every cell in a table must contain an atomic value (e.g., string, date, number), meaning that cell content cannot be composite or multi-valued. This principle, known as the first normal form in database systems~\citep{kiferdatabase}, ensures that cell values can be smoothly queried and updated without introducing anomalies.

The process of value normalization involves several critical steps to ensure data consistency and accuracy. First, we focus on value splitting and extraction, identifying and splitting all composite columns. This may involve adding new columns as necessary while ensuring that no existing columns are deleted. Next, we standardize date and numerical values to a uniform format, paying special attention to any additional strings such as currency symbols, units or comma that may accompany numerical values. Additionally, we normalize all ``N/A'' and blank values to NULL to maintain consistency throughout the dataset. In SQL, null values signify an attribute value that is not available or missing, and they are treated differently than any other values. SQL engines recognize the semantics of null values and consider this when processing queries. For columns containing value ranges, such as ``2010/11'' or ``2015-2018'', we split these into two separate columns to facilitate clearer data interpretation and processing.

An example of value normalization is shown in Figure \ref{fig:NormTab-example}. The original table presents date columns with dates in textual format, a result column combining match outcomes with scores, and an attendance column where numbers are written with commas. The value representation in the original table is more readable for humans; however, this format poses challenges for symbolic programs to process.
Our normalization process converts the date to the ``YYYY-MM-DD'' format and attendance values to a pure numerical format by removing commas. Additionally, NormTab splits the composite result column into two separate columns: ``result\_type'' and ``result\_score'', thereby organizing the data more effectively for analysis. This standardization is crucial for maintaining data integrity across the table. \\

\textbf{Structural Normalization:} 
Tables can be organized either row-oriented or column-oriented. In a row-oriented table, each row typically represents an entity or a relationship between entities, while each column describes an attribute of the entity or relationship. Column-oriented tables, on the other hand, are stored in a transposed fashion. Most traditional databases store data in a row-oriented format, which is well-supported across relational databases.

Our structure normalization primarily focuses on addressing structural differences between tables to enhance their usability for reasoning tasks. Initially, we carefully examine the table structure to determine if the first row resembles a header, indicating the table is row-oriented and requires no structural changes. However, if the first column appears to serve as the header, we transpose the table to normalize its structure, ensuring that the layout aligns with our adopted tabular format. Additionally, web tables sometimes include aggregated rows or columns, which can pose challenges if specific rows or columns need aggregation to answer a query.  We handle these aggregated rows by disregarding any information present in the last row that pertains to aggregated data, such as ``total'', ``sum'', or ``average''. This step prevents redundant or misleading data from affecting subsequent analyses and ensures that the table remains clean and focused on the relevant data points.

\subsection{Normalization Approach: NormTab}
As depicted in Figure \ref{fig:NormTab}, our methodology for normalizing web tables involves two distinct approaches to leverage the capabilities of LLMs for enhancing symbolic reasoning and query capabilities. 

\textbf{Entire Table Normalization (NormTab-Basic):} In the first approach, we provide the LLM with the entire table along with specific instructions for cleaning and normalizing. The LLM reads the table and the instructions, then returns a cleaned and normalized version of the table. However, we observed that many web tables contain portions already in a well-structured form, with only a few columns requiring normalization. To optimize this process, we developed a modified approach. \\

\textbf{Targeted Normalization(NormTab-Targeted):} To improve efficiency, we developed a modified approach that targets only the portions of the table requiring normalization. Our analysis of web tables revealed that often only a few columns need the normalization process. This realization led to a more optimized methodology. In this more refined approach, we first ask the LLM to identify which columns require normalization and cleaning, based on the table metadata (such as column headers and titles) and a few sample rows. Once these columns are identified, we split the original table into two subtables: one that requires normalization and cleaning, and one that is already normalized and clean. We then send only the subtable that needs normalization to the LLM along with the instructions. The LLM processes this subtable and returns a cleaned and normalized version. After normalization, we merge the normalized subtable with the already clean portion of the table. This approach not only improves the efficiency of the normalization task by reducing the amount of data sent to the LLM but also ensures that the resulting table is in a consistent and accurate format suitable for subsequent reasoning and querying tasks.

Following this, we analyze the overall structure of the merged table. With the assistance of the LLM, we determine whether the table needs to be transposed based on its layout. If needed, table transposition is performed outside of the LLM. Additionally, we check if the last row contains summarized or aggregated values and if so, NormTab ignore this row. This selective column normalization method reduces the workload on the LLM, enhances efficiency, and ensures that only the necessary parts of the table are processed, thereby preserving the integrity of already structured data. \\

\section{Experimental Setup}
\subsection{Dataset}
We conduct experimental evaluations using two challenging web table datasets: WikiTableQuestion (WikiTQ)\citep{pasupat-liang-2015-compositional} and TabFact \citep{Chen2020TabFact:}. These datasets are specifically curated to test the reasoning capabilities of models on complex tabular data. WikiTQ comprises tables extracted from Wikipedia along with corresponding natural language questions, while TabFact consists of tables sourced from Wikipedia paired with textual facts. These datasets provide a diverse range of table structures and content, allowing us to thoroughly evaluate the performance of NormTab in enhancing table reasoning tasks. 

The WikiTQ standard test set comprises 416 unique tables and 4,344 samples, while the TabFact standard test set includes 298 unique tables with 2,003 samples. By utilizing these datasets, we aim to demonstrate the effectiveness of web table normalization in improving the symbolic reasoning performance of LLMs, thereby highlighting the importance of addressing the challenges posed by web table irregularities. 

\subsection{Baselines and Evaluation Metrics}
We compare our approach with several robust baseline methods, including TableCoT \citep{chen-2023-large}, BINDER \citep{cheng2022binding}, DATER \citep{10.1145/3539618.3591708}, StructGPT \citep{jiang-etal-2023-structgpt}, ReAcTable \citep{zhang2023reactable}, Rethinking-Tab-Data \citep{liu2023rethinking}, TabSQLify \citep{nahid2024tabsqlify}, and Chain-of-Table \citep{wang2024chain}.

For the WikiTQ dataset, exact match (EM) accuracy was used to check if the predicted answers matched the correct ones. To address varying text formats, a pre-matching check using LLMs was incorporated \cite{cheng2022binding}. The accuracy for TabFact was assessed using binary classification accuracy.

\subsection{Implementation}
We utilized gpt-3.5-turbo-0125 as the Language Model which supports 16k context window. 
We were inspired by the prompting style from \citep{liu2023rethinking, nahid2024tabsqlify} in our implementation of NormTab.
To compare performance, we employ few-shot in-context learning. This involves supplying the LLM with the table title, table header, question, and three example rows of the table, along with the question, to generate an SQL query. The SQL query is then executed on the table to obtain the answer. Further details can be found in Appendix \ref{appendix: implementation}, and all our code and prompts are available at \url{https://github.com/mahadi-nahid/NormTab}.


\section{Results}

In this section, we analyzed the performance of NormTab. To evaluate its impact, we conducted few-shot in-context learning experiments to generate SQL queries for answering specific questions. First, we performed experiments on unnormalized tables without any modifications. Then, we compared the performance on normalized tables. Additionally, we reported the performance of different normalization processes.

\subsection{Results on Downstream Tasks}
Table \ref{tab: normtab-performacne-wtq} and Table \ref{tab: normtab-performacne-tf} presents a comparison between the performance of NormTab and the other baselines on WikiTQ and TabFact datasets. 

In the WikiTQ dataset, the results showed that after applying the targeted version of NormTab, we achieved 61.2\% accuracy, surpassing the performance of other baseline models. The targeted NormTab approach performs slightly better than the basic version, where the entire table is passed to the LLMs. This suggests that LLMs may be more effective at normalization tasks when dealing with targeted smaller tables. Additionally, we gained about 10\% improvement compared to the Text-to-SQL \citep{rajkumar2022evaluating} model and SQL (gpt-3.5-turbo) model. Notably, Rethinking-Tab-Data \cite{liu2023rethinking} achieved an accuracy of 56.87\% by addressing structural varience using LLMs and a Python agent. Chain-of-Table \citep{wang2024chain} employed an iterative sequence of operations to tailor complex tables to specific questions, achieving 59.94\% accuracy. However, these and other baseline models are question-dependent. In contrast, our model adopts a straightforward and simple approach: it normalizes the table only once, irrespective of the question, enabling answers to be derived from the normalized table using program-aided symbolic reasoning. 

\begin{table}[h]
\centering
\begin{tabular}{lc}
\hline
\textbf{Model}    & \textbf{Acc (\%)} \\ \hline \hline
TableCoT \citep{chen-2023-large} & 52.40     \\ \hline
BINDER &  56.74  \\ \hline
DATER & 52.80 \\ \hline
ReAcTable & 52.40 \\ \hline 
Rethinking-Tab-Data & 56.87 \\ \hline
Chain-of-Table & 59.94 \\ \hline \hline
Text-to-SQL \citep{rajkumar2022evaluating} & 52.90 \\ \hline
Text-to-SQL (gpt-3.5-turbo) & 51.30 \\ \hline
\textbf{NormTab\textsubscript{\textit{Basic}} + SQL (ours)} & \textbf{60.80}    \\ \hline
\textbf{NormTab\textsubscript{\textit{Targeted}} + SQL (ours)} & \textbf{61.20}    \\ \hline
\end{tabular}
\caption{Performance comparison of NormTab on WikiTQ dataset. The results clearly demonstrate that NormTab significantly surpasses other models in accuracy when employing symbolic reasoning.}
\label{tab: normtab-performacne-wtq}
\end{table}

In Table \ref{tab: normtab-performacne-tf}, we can observe a similar performance enhancement compared to the original table in table-based fact verification tasks. We achieved approximately a 6\% performance improvement compared to the results of Text-to-SQL on the original table. It is worth noting that table-based fact verification differs from table-based question answering tasks. Generating a SQL query to verify a fact is more complex than simply retrieving an answer from the table. Although other models not employing program-aided symbolic reasoning perform better in this task, these models utilize LLMs for the verification task providing the whole table to the model. Our experimental results show promise for utilizing symbolic reasoning in such scenarios. 

\begin{table}[h]
\centering
\begin{tabular}{lc}
\hline
\textbf{Model}    & \textbf{Acc (\%)} \\ \hline \hline
TableCoT-chatgpt  & 73.10 \\ \hline 
BINDER     & 79.17 \\ \hline 
DATER     & 78.01 \\ \hline 
Chain-of-Table  & 80.20 \\ \hline 
ReAcTable & 73.10 \\ \hline \hline
Text-to-SQL \citep{rajkumar2022evaluating}  & 64.71 \\ \hline
Text-to-SQL (gpt-3.5-turbo) & 62.32 \\ \hline
\textbf{NormTab\textsubscript{\textit{Basic}} + SQL (ours)}  & \textbf{67.10}    \\ \hline
\textbf{NormTab\textsubscript{\textit{Targeted}} + SQL (ours)}  & \textbf{68.90}    \\ \hline
\end{tabular}
\caption{Performance comparison of NormTab on TabFact dataset with other models.}
\label{tab: normtab-performacne-tf}
\end{table}


In this study, we also evaluated the effectiveness of our method using Gemini-1.5-flash and GPT-4-turbo. The results showed a improvement on the WikiTableQuestions dataset, demonstrating that our model's performance is not heavily dependent on specific language models (see Table \ref{tab: normtab-performacne-wtq-gemini-gpt4}). 

\begin{table}[h]
\centering
\begin{tabular}{lc}
\hline
\textbf{Model}    & \textbf{Acc (WTQ)} \\ \hline \hline
NormTab\textsubscript{\textit{Basic}} (gemini-1.5-flash)  & 61.36    \\ \hline
NormTab\textsubscript{\textit{Targeted}} (gemini-1.5-flash)  & 61.24   \\ \hline
NormTab\textsubscript{\textit{Basic}} (gpt-4-turbo)  & 61.57    \\ \hline
NormTab\textsubscript{\textit{Targeted}} (gpt-4-turbo)  & \textbf{62.28}    \\ \hline
\end{tabular}
\caption{Performance of NormTab on WikiTQ dataset using Gemini-1.5-flash and GPT-4-turbo. }
\label{tab: normtab-performacne-wtq-gemini-gpt4}
\end{table}


\subsection{NormTab Evaluation}
To assess the accuracy of various normalization operations, we evaluated the performance on 100 tables, with 50 tables from each dataset, WikiTQ and TabFact. Table \ref{tab:normtab-acc} summarizes the accuracy of different normalization processes. 
NormTab demonstrated strong performance in normalizing dates and numbers, detecting transposition requirements, and handling aggregated summaries in the last row effectively. However, NormTab faced difficulties in extracting and cleaning values in certain critical tables where value extraction from the original table was particularly challenging. The column selection accuracy indicates that LLMs can be very effective in identifying columns where values are not in the proper format. However, the accuracy of splitting columns was low. Additional errors included managing value cleaning and handling "n/a" values. Although these tasks are challenging, the performance in these areas shows the potential for utilizing LLMs to address these tasks effectively.

\begin{table}[h]
\centering
\begin{tabular}{lr}
\hline
\textbf{Type}           & \textbf{Accuracy}\\ \hline
Columns Selection & 91.0\% \\ \hline 
Transpose Detection           & 97.0\%     \\ \hline
Last Row Aggregation & 100.0\%      \\ \hline
Split Column         & 87.0\%     \\ \hline
Date and Number      & 100.0\%      \\ \hline
N/A value            & 93.0\%     \\ \hline
Value Cleaning       & 82.0\%     \\ \hline
\end{tabular}
\caption{Accuracy of \textit{\textbf{NormTab}} in various types of normalization.}
\label{tab:normtab-acc}
\end{table}

NormTab has shown superior performance compared to several robust models, demonstrating its efficacy in table normalization. A key advantage of NormTab is its use of program-aided symbolic reasoning, which streamlines code generation without requiring the entire table to be passed to the LLM. This enhances efficiency and eliminates dependencies on table size and answer position. With NormTab, only key elements like the title, header, and a few example rows are needed to generate SQL queries and obtain accurate answers. This approach reduces computational overhead while maintaining high accuracy, highlighting its practical utility in various table-based tasks.


Our normalization method, NormTab, can be beneficial for a variety of table reasoning tasks, especially those employing symbolic methods. For the same reason, we have integrated NormTab with a recent table reasoning method TabSQLify \citep{nahid2024tabsqlify}, which utilizes symbolic techniques. Our evaluation reveals that integrating NormTab with TabSQLify leads to a notable 4\% improvement in performance. This demonstrates NormTab's potential to enhance other symbolic frameworks by serving as an effective preprocessing step. The results is summarized in Table \ref{tab: normtab-tabsqlify}. 

\begin{table}[h]
\centering
\begin{tabular}{lc}
\hline
\textbf{Model}    & \textbf{Acc (WTQ)} \\ \hline \hline
TabSQLify   & 64.7    \\ \hline
NormTab+TabSQLify  & \textbf{68.63}    \\ \hline
\end{tabular}
\caption{Performance of TabSQLify integrated with NormTab on the WikiTableQuestions dataset (gpt-3.5-turbo).}
\label{tab: normtab-tabsqlify}
\end{table}

In our work, each table is normalized once, as a preprocessing step, irrespective of the number and the type of questions asked. Our approach does not depend on the specific question or task. While using Self-Consistency \citep{wang2023selfconsistency, liu2023rethinking} might seem beneficial, it requires generating multiple responses per question, which can increase costs and reduce efficiency. For instance, if we need answers to 10 questions and apply Self-Consistency with 6 paths, we end up with 60 samples (10 questions × 6 paths). Additionally, we need to send the entire table with each question, resulting in a higher number of tokens. In contrast, our method requires just a few prompts to normalize the table initially. After normalization, we do not need to pass the entire table to generate the SQL query for each question; only table metadata such as the table title, column names, and a few example rows, is needed. This one-time preprossessing step significantly reduces the overall number of tokens passed to the LLM, potentially lowering costs compared to using Self-Consistency. 

\subsection{Analysis}

We conducted a detailed analysis of the impact of NormTab on the WikiTQ dataset. Table \ref{tab:category} shows that in \textbf{67\%} of cases (Category A), performance improved after applying NormTab. In 24\% of cases (Category B), performance remained unchanged, indicating no improvement. Additionally, in 9\% of cases (Category C), performance actually decreased. The detailed experimental findings are summarized in Table \ref{tab:result_analysis}.

\begin{table}[h]
\small
\begin{tabular}{clc}
\hline
\textbf{Categories} & \textbf{Description}        & \textbf{\% of Tables} \\ \hline \hline 
A          & \begin{tabular}[c]{@{}l@{}}Where performance\\ enhanced after \\ applying NormTab\end{tabular}       & \textbf{67\%}   \\ \hline
B          & \begin{tabular}[c]{@{}l@{}}Where no change in \\ performance after \\ applying NormTab\end{tabular}  & 24\%    \\ \hline
C          & \begin{tabular}[c]{@{}l@{}}Where the performance \\ decreased after \\ applying NormTab\end{tabular} & 9\%      \\ \hline
\end{tabular}
\caption{Categories of tables on WikiTQ test dataset.}
\label{tab:category}
\end{table}

\begin{table}[h]
\centering
\small
\begin{tabular}{lccc}
\hline
   \textbf{ - }     & \textbf{Tables (A)} & \textbf{Tables (B,C)} & \textbf{Overall (A,B,C)} \\ \hline \hline
Original & 46.28\%               & 59.62\%                  & 51.30\%             \\ \hline
NormTab  & \textbf{62.55\%}               & 56.76\%                    & \textbf{61.20\%}             \\ \hline
Change   & \textbf{+16.27}              & -2.86                     & \textbf{+9.9}             \\ \hline
\end{tabular}
\caption{Result breakdown on WikiTQ dataset.}
\label{tab:result_analysis}
\end{table}

Table \ref{tab:result_analysis} demonstrates that NormTab can improve overall performance by 9.9\%. Notably, in Category A, we observed a substantial enhancement of 16.27\%. However, Categories B and C saw a slight decline in performance due to highly complex table values and structures. 


The basic NormTab approach involves only one LLM call, but it requires passing the entire table to the language model as a prompt. This means the LLM must process a larger number of tokens, which can impact the overall normalization performance. Research indicates that more tokens can increase the likelihood of hallucination and can be more costly \citep{chen-2023-large,10.1145/3539618.3591708, 10.1145/3571730}. In contrast, the targeted NormTab approach first filters out the parts of the table that are already well-formatted, thereby reducing the number of tokens sent to the LLM by focusing only on the columns that need normalization. For example, consider a table with 15 rows and 8 columns, resulting in a total of 15 * 8 = 120 table cells. In the basic NormTab approach, all 120 cells are sent to the language model along with the instructions. However, if we identify that only 5 out of 8 columns require normalization, we only need to send 15 * 5 = 75 table cells. This reduction translates to 45 fewer table cells, which is a 37.5\% reduction in table size. Table \ref{tab:reduction} illustrates that we can achieve a 72\% reduction in table size for both datasets by employing the targeted NormTab approach, which is quite substantial.

\begin{table}[h]
\centering
\small
\begin{tabular}{lccc}
\hline
    \textbf{Dataset}     & \textbf{\begin{tabular}[c]{@{}c@{}}NormTab\\(Basic)\end{tabular}} & \textbf{\begin{tabular}[c]{@{}c@{}}NormTab\\(Targeted)\end{tabular}} & \textbf{Reduction} \\ \hline \hline
WikiTQ & 152.26         & 41.82                 & \textbf{72.53\%}             \\ \hline
TabFact  & 106.19               & 29.11	                   & \textbf{72.58\%}             \\ \hline
\end{tabular}
\caption{Table cell reduction in NormTab-Targeted compared to NormTab-Basic}
\label{tab:reduction}
\end{table}

Although the targeted NormTab approach requires additional LLM calls for tasks such as column selection and transposition detection, the total number of tokens processed is still lower than in the basic NormTab approach. While the basic approach may be suitable for smaller tables, the reduction in table size is crucial for normalizing larger tables effectively. The targeted strategy involves multiple queries, but it refines the normalization process by concentrating on specific subtasks, which may help reduce hallucination and errors. While the performance improvement appears marginal, the significant token size reduction makes the targeted NormTab approach highly beneficial for larger tables.

\section{Conclusion}
In conclusion, our study introduces NormTab, a framework aimed at enhancing LLMs' performance on tabular data by normalizing web tables. Through our investigation, we have shown the significance of web table normalization in overcoming challenges such as mixed values and structural variance. By leveraging LLMs' textual understanding in data cleaning and normalization, NormTab improves table reasoning. Our experiments on challenging datasets demonstrate its effectiveness. Our work contributes to advancing techniques for LLMs in handling tabular data, emphasizing the importance of addressing web table challenges for improved performance. Further research can explore additional normalization strategies and extend NormTab's applicability across various domains. This would establish a robust foundation for a wide range of downstream tasks involving table reasoning. 

\section*{Limitations}
Despite the advancements brought by NormTab, there are several limitations. First, while our framework significantly enhances the symbolic reasoning capabilities of LLMs on tabular data, there remains room for improvement in the normalization process, particularly with more complex table structures. Additionally, for larger tables, LLMs may sometimes produce hallucinated results, leading to inaccuracies in the normalized output, indicating a need for better handling of extensive datasets. Moreover, when working with tables containing highly noisy data, LLMs often struggle to clean and normalize the information effectively, and may generate output in an incorrect format, making it challenging to parse. The presence of excessive noise and inconsistencies can hinder the normalization process and negatively impact overall performance. Addressing these limitations is crucial for further enhancing the robustness and reliability of NormTab. As we measure the accuracy using the results obtained from LLM based Text-to-SQL model, it is important to note that some questions in the dataset may not directly map to SQL queries which may affect the performance. 

\section*{Acknowledgments}
We extend our sincere gratitude to all reviewers for their invaluable feedback, insightful suggestions, and positive remarks about our work. This research has been supported by the Natural Sciences and Engineering Research Council of Canada. Also, Md Mahadi Hasan Nahid was supported by the Alberta Innovates Graduate Student Scholarship.

\section*{Ethical Considerations}
The datasets used in this study are accessible through the peer-reviewed articles cited in the references section. Additionally, our source code is openly available for future research under the MIT License. It is important to mention that our framework relies on GPT-3.5-turbo, which may inherit ethical concerns associated with GPT models. These concerns include the potential for generating responses to toxic content or displaying biased behavior.

\bibliography{custom}

\appendix

\section{Web Tables vs Regular DB Table}
\label{appendix: webvsdb-table}

There are some key differences between a web table and relational database table. Web tables and relational database tables exhibit distinct characteristics, each with its own advantages and challenges. Web tables, often sourced from online sources such as websites and spreadsheets, tend to be diverse in structure and content. They may contain varying formats, including column-based, row-based, and aggregated summaries, with irregularities such as mixed values and noisy data being common. Additionally, web tables frequently present challenges in terms of structural variance, as they may lack standardized schema and exhibit inconsistencies in data organization. In contrast, relational database tables (RDBMS) adhere to a structured schema. These tables are typically designed for efficient storage and retrieval of data, with clear relationships defined between entities through keys and constraints. Relational database tables offer advantages in terms of data integrity, consistency, and scalability, as they enforce normalization principles and allow for efficient querying through SQL. 

However, Web Tables may lack the flexibility and adaptability required to handle the diverse and unstructured nature of web data. While relational database tables excel in maintaining structured data integrity, web tables present challenges related to variability and noise, necessitating specialized techniques for effective processing and analysis.

\section{Comparison with Other Models}
\label{appendix: comparison}

Chain-of-Table \citep{wang2024chain} enhances reasoning on tabular data by iteratively transforming and evolving table structures through a series of reasoning steps, including row/column selection and cell splitting. Their method uses in-context learning to guide LLMs in generating operations and updating the table, forming a reasoning chain specific to tabular data. \citet{liu2023rethinking} explore LLM capabilities in interpreting and reasoning over tabular data, emphasizing robustness to structural changes and comparing textual and symbolic reasoning. They find that structural variations can significantly degrade performance, especially in symbolic reasoning tasks, and propose table structure normalization through transposition to mitigate this issue. Their study concludes that while textual reasoning slightly outperforms symbolic reasoning, both have distinct strengths depending on the task.

The ReAcTable model \citep{zhang2023reactable} follows the ReAct paradigm, incorporating step-by-step reasoning, external tool-based code execution, intermediate table generation, and majority voting to process tabular data. Recent approaches leverage LLM capabilities for these tasks. For instance, \citet{narayan2022can} demonstrated LLM effectiveness in identifying errors and imputing missing data, enhancing the data wrangling process. Integrating LLMs can significantly improve the efficiency and accuracy of preparing data for analysis, streamlining and automating many aspects of data wrangling. Operations like normalizing numbers and dates can be incorporated into data processing workflows to aid subsequent analysis \citep{nahid2024tabsqlify, cheng2022binding}.

Existing models and methods typically rely on multi-step frameworks where LLMs select actions, such as adding columns, and additional scripts process values based on specific questions \citep{liu2023rethinking, wang2024chain, zhang2023reactable}. However, these approaches are question-dependent and do not comprehensively address the root issue of table normalization. NormTab differs by focusing on normalizing tables once, regardless of the question, allowing answers to be derived from the normalized table using program-aided symbolic reasoning. This approach reduces dependencies on table size and answer position, enhancing efficiency and versatility in table reasoning tasks.

\section{Implementation Settings}
\label{appendix: implementation}

The dataset we used contains only the gold answers and lacks the original SQL queries needed to extract these answers. Additionally, the dataset does not include normalized or cleaned versions of the tables. Data contamination can indeed impact methods where the question and table are provided to the LLM, which might then rely on knowing the gold answer directly. In our approach, however, we focus on converting tables into a normalized format. Since the dataset does not provide such cleaned versions, data contamination does not affect our method. Moreover, our approach generates SQL queries based on the question and table metadata, rather than relying on pre-existing gold SQL queries. We obtain the answers by executing these generated SQL queries. Therefore, data contamination does not impact our method.

Our implementation of NormTab was inspired by the prompting techniques used in \citep{liu2023rethinking, nahid2024tabsqlify}. We configured the in-context learning hyperparameters for gpt-3.5-turbo-0125 according to the specifications outlined in Table \ref{tab:hyper-parameters1}.

\label{appendix:a}
\begin{table}[h]
\centering
\small
\begin{tabular}{lccc}
\hline
\textbf{Parameter}             & \begin{tabular}{c} \textbf{Coll} \\ \textbf{Selection} 
\end{tabular}          & \begin{tabular}{c} \textbf{Transpose} \\ \textbf{Detection} \end{tabular}   & \textbf{NormTab} \\ \hline \hline
temperature           & 0.3             & 0.3           & 0.7       \\ 
top\_p                & 1               & 1             & 1          \\
sample\_n             & 1               & 1             & 1    \\ 
max\_tokens           & 100             & 100           & 4500    \\ 
num\_shots            & 6              & 1           & 1   \\ \hline

\end{tabular}
\caption{The hyper-parameters we set in NormTab}
\label{tab:hyper-parameters1}
\end{table}

\section{Example Prompts}
\label{appendix: prompts}

The prompt used in NormTab is described in the following Figures. 

\begin{figure*}[h]
    \centering
    \resizebox{\textwidth}{!}{
    \includegraphics{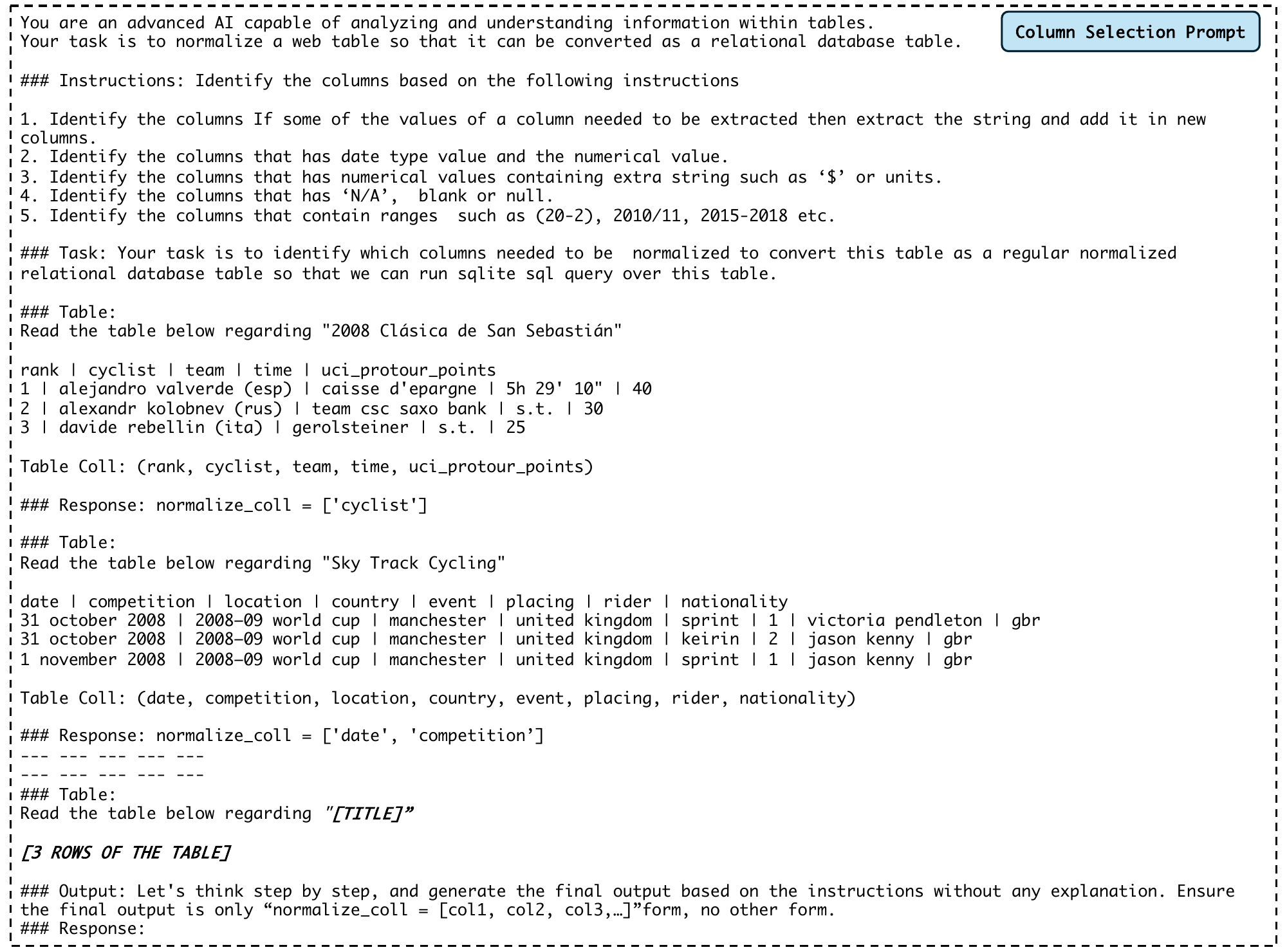}
    }
    \caption{Column Selection prompt.}
    \label{fig:NormTab-col-prompt}
\end{figure*}

\begin{figure*}[h]
    \centering
    \resizebox{\textwidth}{!}{
    \includegraphics{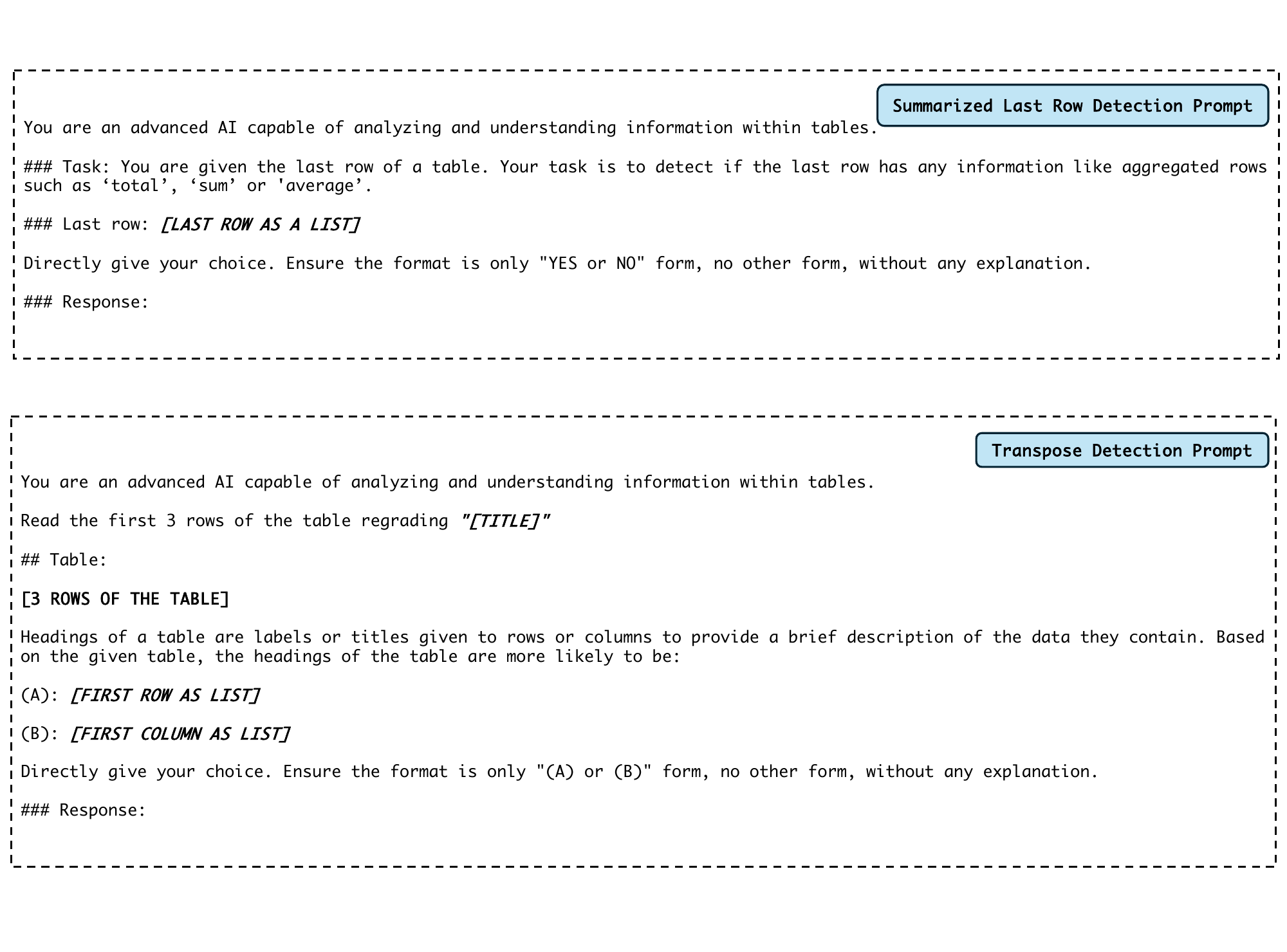}
    }
    \caption{Summarized last row detection and transpose detection prompt.}
    \label{fig:NormTab-last-row-transpose-prompt}
\end{figure*}

\begin{figure*}[h]
    \centering
    \resizebox{\textwidth}{!}{
    \includegraphics{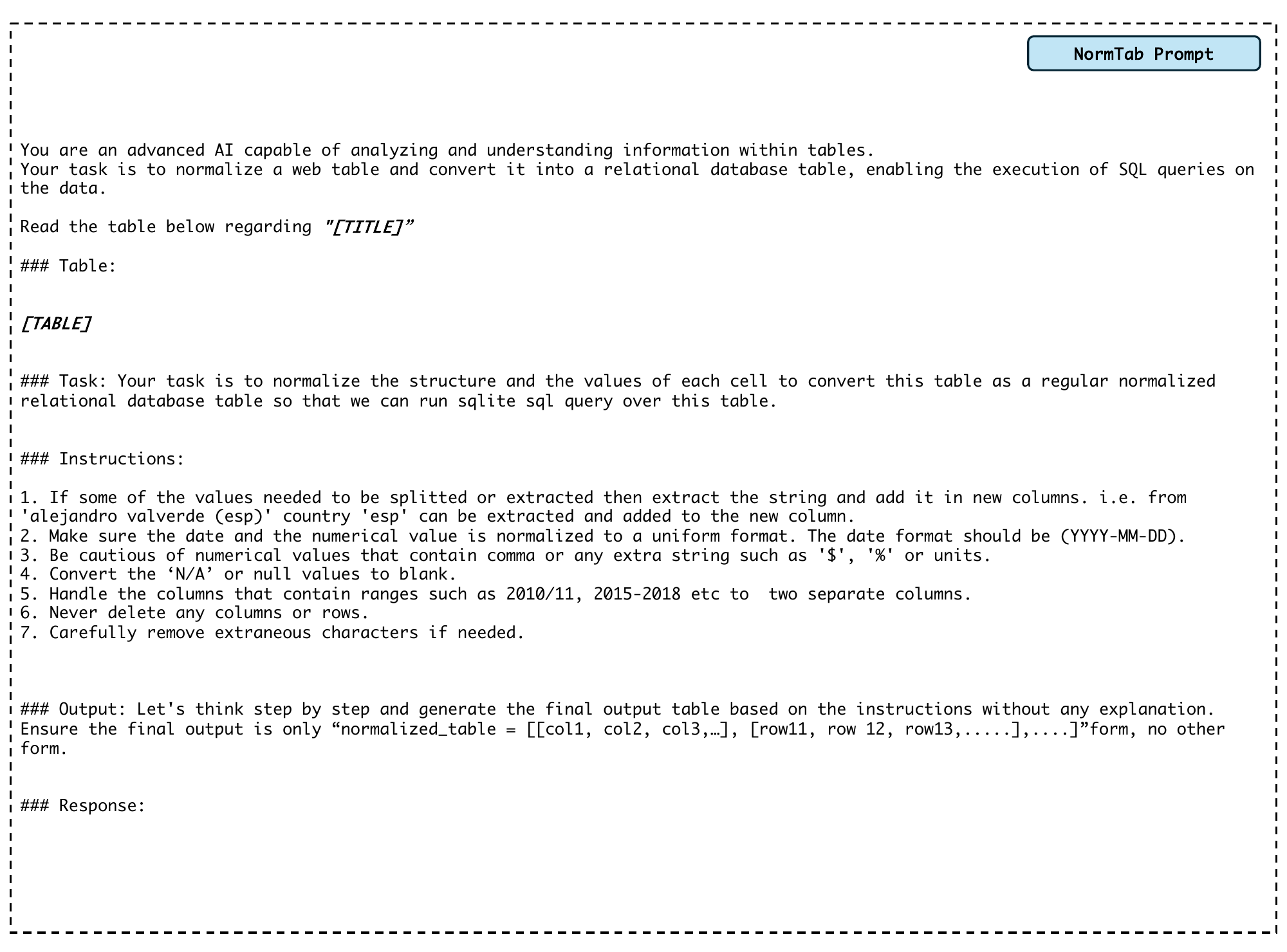}
    }
    \caption{NormTab Instruction prompt.}
    \label{fig:NormTab-prompt}
\end{figure*}

\begin{figure*}[h]
    \centering
    \resizebox{\textwidth}{!}{
    \includegraphics{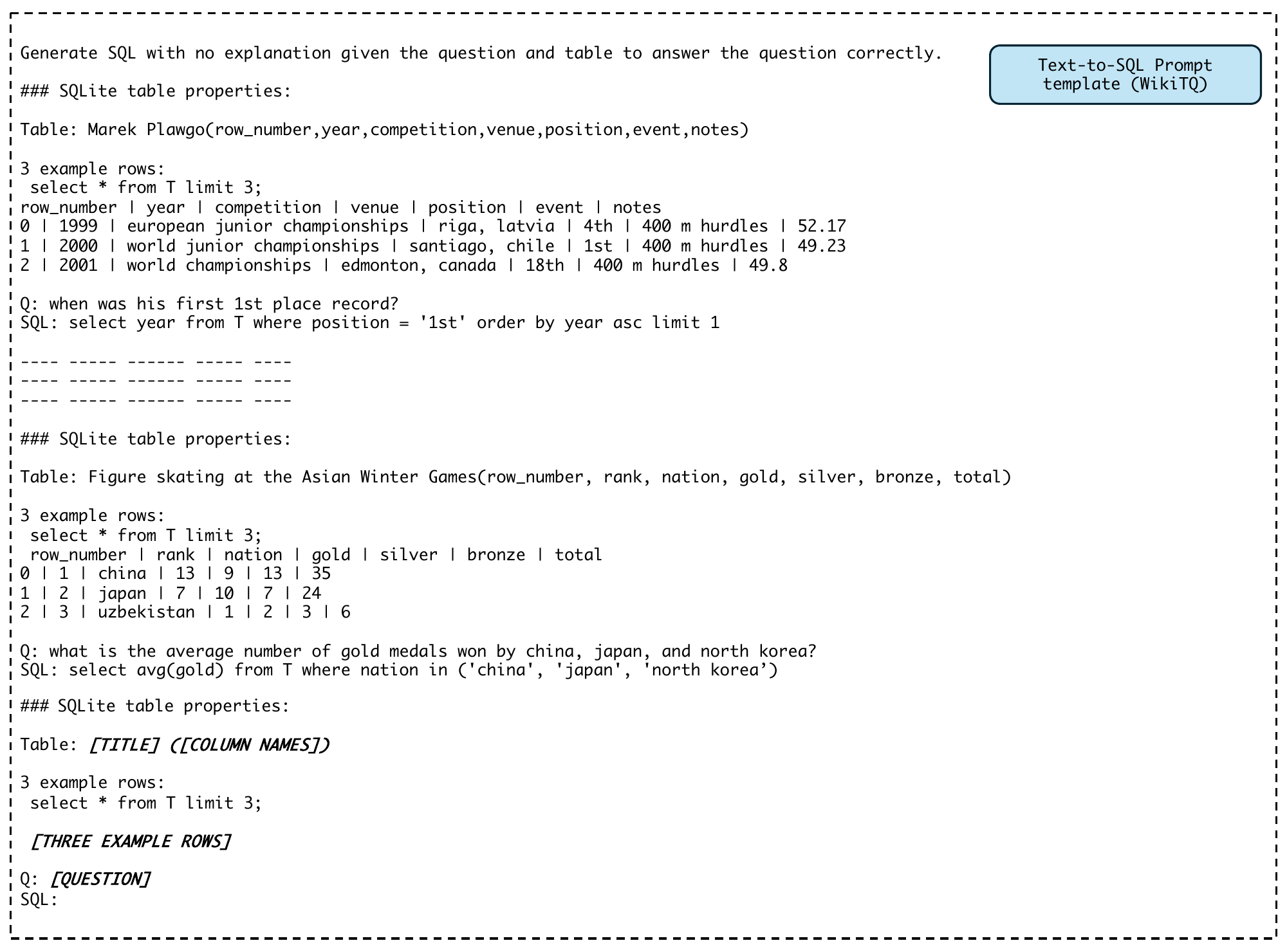}
    }
    \caption{Text-to-SQL prompt template for the Table QA Task on the WikiTQ dataset.}
    \label{fig:NormTab-sql-wtq}
\end{figure*}

\begin{figure*}[h]
    \centering
    \resizebox{\textwidth}{!}{
    \includegraphics{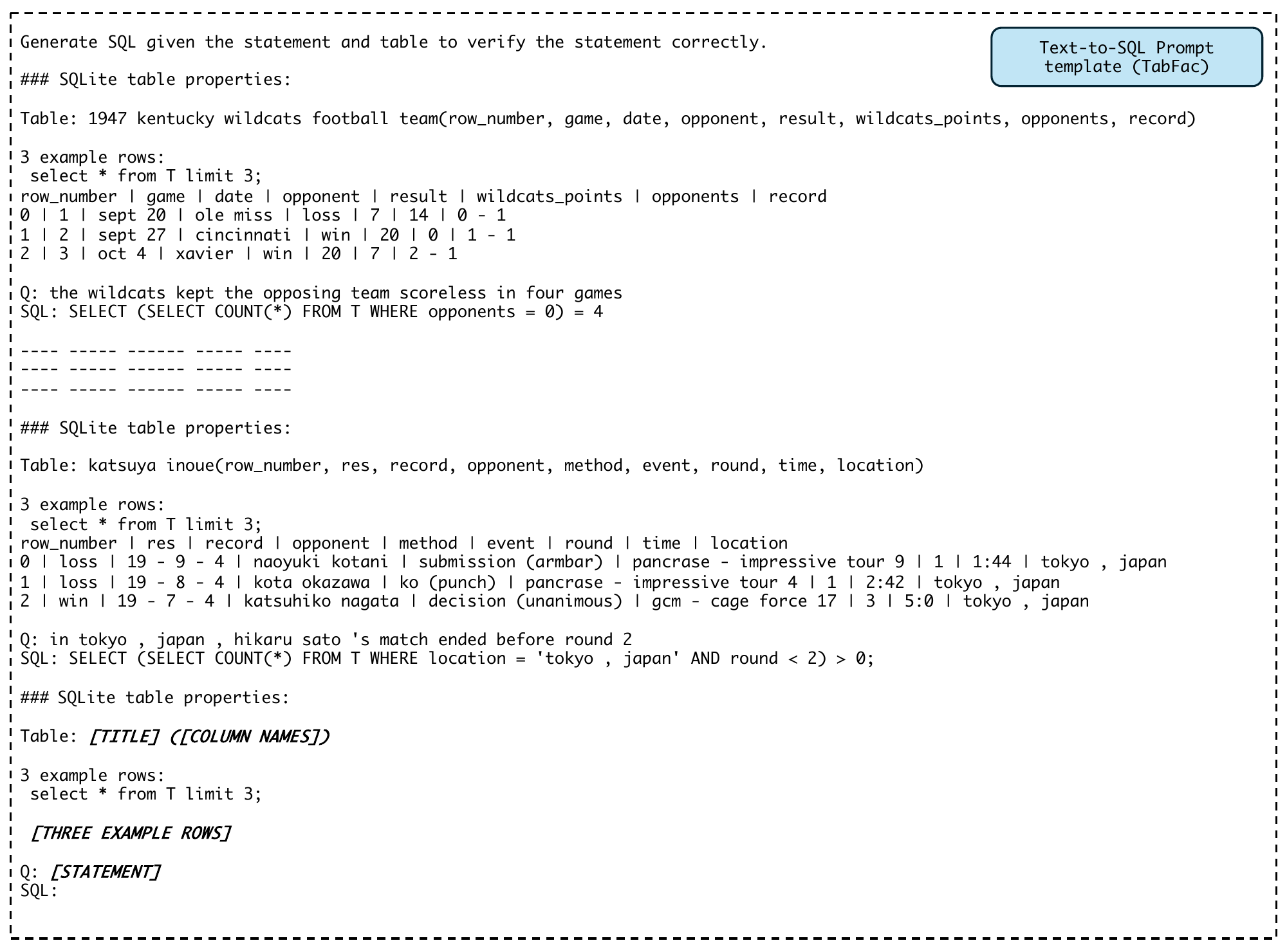}
    }
    \caption{Text-to-SQL prompt template for the Table-based Fact Verification Task on the TabFact dataset.}
    \label{fig:NormTab-sql-tf}
\end{figure*}

\end{document}